%% file: ms.tex
\documentclass{article}

\usepackage[preprint]{neurips_2025}

\usepackage[utf8]{inputenc} 
\usepackage[T1]{fontenc}    
\usepackage{hyperref}       
\usepackage{url}            
\usepackage{booktabs}       
\usepackage{amsfonts}       
\usepackage{nicefrac}       
\usepackage{microtype}      
\usepackage{xcolor}         
\usepackage{xspace}
\usepackage{nolbreaks}
\usepackage{graphicx}
\usepackage{cleveref}
\usepackage{mathabx}
\usepackage{subcaption}
\usepackage{enumitem}
\usepackage{authblk}

\newif\ifshowcomments
\showcommentsfalse  

\ifshowcomments
    \newcommand{\alvin}[1]{{\color{teal} Alvin: [{#1}]}}
    \newcommand{\linyuan}[1]{{\color{orange} Linyuan: [{#1}]}}
    \newcommand{\mostafa}[1]{{\color{purple} Mostafa: [{#1}]}}
    \newcommand{\sida}[1]{{\color{blue} Sida: [{#1}]}}
\else
    \newcommand{\alvin}[1]{}
    \newcommand{\linyuan}[1]{}
    \newcommand{\mostafa}[1]{}
    \newcommand{\sida}[1]{}
\fi

\title{Structure-Aware Fill-in-the-Middle Pretraining for Code}

\author{%
  \textbf{Linyuan Gong}$^{1}$, 
  \textbf{Alvin Cheung}$^{1}$, 
  \textbf{Mostafa Elhoushi}, 
  \textbf{Sida Wang}$^{2}$
}
\affil{\small{$^{1}$UC Berkeley, $^{2}$FAIR at Meta} \\
\texttt{\{gly, akcheung\}@berkeley.edu, m.elhoushi@ieee.org, sida@meta.com}}

\begin{document}

\maketitle

\input{contents/abstract}


\input{contents/introduction}

\input{contents/related_work}

\input{contents/method}

\input{contents/real_fim_eval_benchmark}

\input{contents/experimental_setup}

\input{contents/evaluation_results}

\input{contents/limitations}

\input{contents/conclusion}

\begin{ack}
\input{contents/acknowledgement}
\end{ack}

\small
\bibliographystyle{assets/plainnat}
\bibliography{references}
\normalsize

\clearpage
\appendix
\section{Appendix}
\input{contents/appendix_l2r_prompt}

\input{contents/appendix_stats_real_fim_eval}

\input{contents/appendix_hyperparameters}

\input{contents/appendix_method_analysis}

\input{contents/appendix_broader_impact}

\if@neuripsfinal
\input{contents/paper_checklist}
\fi

\end{document}

%% file: contents/abstract.tex
\begin{abstract}
Fill-in-the-Middle (FIM) is a common pretraining method for code LLMs, where models complete code segments given surrounding context. However, existing LLMs treat code as plain text and mask random character spans.
We propose and evaluate AST-FIM, a pretraining strategy that leverages Abstract Syntax Trees (ASTs) to mask complete syntactic structures at scale, ensuring coherent
training examples better aligned with universal code structures and common code editing patterns such as blocks, expressions, or functions. To evaluate real-world fill-in-the-middle (FIM) programming tasks, we introduce Real-FIM-Eval, a benchmark derived from 30,000+ GitHub commits across 12 languages.
On infilling tasks, experiments on 1B and 8B parameter models show that AST-FIM is particularly beneficial for real-world code editing as it outperforms standard random-character FIM by
up to 5 pts on standard FIM benchmarks. 
Our code is publicly available at {\ttfamily \url{https://github.com/gonglinyuan/ast_fim}}.

\end{abstract}

%% file: contents/introduction.tex
\section{Introduction}\label{sec:introduction}


Large Language Models (LLMs) trained on diverse, internet-scale datasets have shown remarkable success across various domains, especially code-related applications. Code LLMs now power code generation, code completion/editing, test generation and much more. Following early works~\citep{humaneval_fim,incoder} and perhaps due to the importance of code editing, \textit{fill-in-the-middle (FIM)} pretraining has emerged as a defining feature of recent code LLMs such as CodeLlama~\citep{codellama}, CodeGemma~\citep{codegemma}, StarCoder2~\citep{starcoder2}, DeepSeek-Coder~\citep{deepseekcoder}, Qwen2.5-Coder~\citep{qwencoder}, and Codestral~\citep{codestral}. Specifically, FIM trains an autoregressive decoder-only model for \textit{infilling}: generating the middle span using the prefix and suffix.

However, existing FIM pretraining approaches treat code as character sequences, disregarding its formal syntactic structure. While simple and general, \textit{random-character FIM (Rand-FIM)} splits code at random positions and creates a mismatch with real-world code editing patterns. This mismatch  likely leads to decreased performance~\citep{deepseekcoder} and misses a natural opportunity to use the structure of code.

\input{figures/fig_intro}

We propose \textbf{AST-Aware Fill-in-the-Middle (AST-FIM)}, which aligns pretraining with universal code structures and real-world code edits through syntax-aware masking at scale. Real-world code edits---such as adding statements to a code block or adding functions in a class declaration---often involve complete syntactic units. AST-FIM leverages Abstract Syntax Trees (ASTs) to mask entire subtrees (\Cref{fig:fig_intro}), preserving structural coherence while mimicking developer actions. We implement AST-FIM through a \textit{language-agnostic} masking algorithm based on Tree-sitter~\citep{treesitter_2025} outputs, enabling support for 100+ programming languages\footnote{{\ttfamily \url{https://github.com/tree-sitter/tree-sitter/wiki/List-of-parsers}}} at scale without further language specific engineering. This approach maintains the simplicity of FIM pretraining while aligning the pretraining objective with code structure and developer behavior.


To verify that AST‑FIM's syntax‑aware masking translates into better performance in real-world code completion tasks, we need an evaluation benchmark that mirrors everyday developer activity. However, existing benchmarks are noisy and expensive ~\citep{swebench,copilot_arena} or unrepresentative of real code changes~\citep{humaneval_fim, safim}. We therefore propose \textbf{Real‑FIM‑Eval}, sourced from more than 30,000 recent commits from highly active GitHub projects spanning 12 programming languages. The model is tasked to complete code segments inserted or modified in a git commit, conditioned on their surrounding context. Because every example is a real commit diff, Real‑FIM‑Eval provides an unbiased, faithful view of real‑world code completion and editing abilities of LLMs.


\input{figures/fig_plot}

We evaluate AST-FIM and baselines on both standard L2R generation and code infilling benchmarks, including Real-FIM-Eval. Our experiments in \Cref{fig:fig_plot} show that AST-FIM not only achieves strong performance on code infilling tasks but also retains L2R generation capability, outperforming random-character FIM (Rand-FIM) by significant margins. These trends hold consistently across both 1B and 8B parameter scales, highlighting the scalability of our approach. Notably, AST-FIM-8B rivals leading models of comparable size in L2R code generation benchmarks while surpassing all the counterparts in real-world infilling scenarios. By integrating syntax-aware pretraining into decoder-only LLMs, AST-FIM significantly advances the infilling capability of code LLMs without compromising their core L2R and random-infilling capabilities. 

%% file: figures/fig_intro.tex
\begin{figure}[t]
\vskip 0.2in
\begin{center}
\centerline{\includegraphics[scale=0.85]{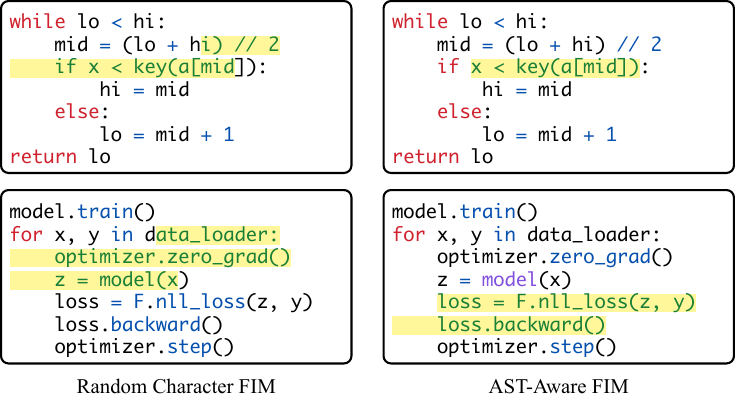}}
\caption{
\textbf{Comparison of masking strategies in Random-Character FIM (Rand-FIM) and our proposed AST-Aware FIM (AST-FIM)} in two examples. The highlighted code is the masked part for FIM training. \textbf{Left:} Rand-FIM treats code as a character sequence, masking a random span.  \textbf{Right:} AST-FIM respects code structure by masking complete subtrees. This syntax-aware masking aligns more closely with typical developer-code interactions.}\label{fig:fig_intro}
\end{center}
\vskip -0.2in
\end{figure}

%% file: figures/fig_plot.tex
\begin{figure}[t]
\vskip 0.2in
\begin{center}
\centerline{\includegraphics[width=0.95\textwidth]{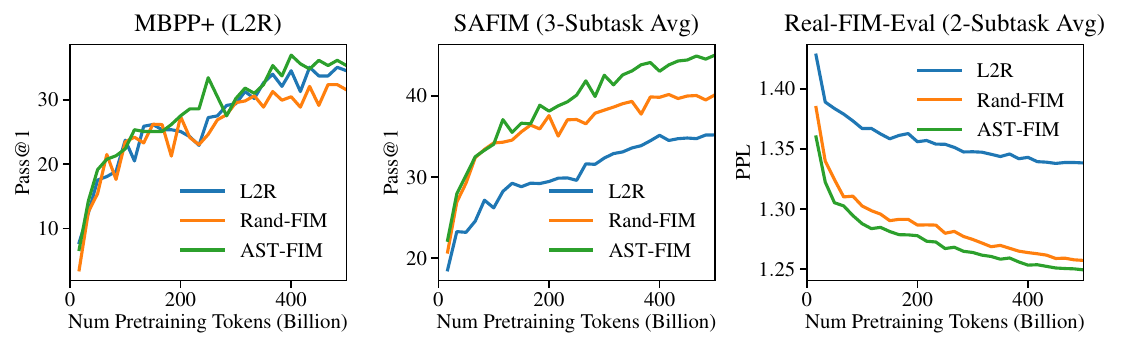}}
\caption{
\textbf{Performance of 1B parameter model during different types of pretraining, checkpointed every 4000 steps (16.7B tokens).}
\textbf{Left:} Pass@1 of MBPP+, a left-to-right task (higher is better).
\textbf{Middle:} Average pass@1 of SAFIM-Algorithm, SAFIM-Control, and SAFIM-API (higher is better). 
\textbf{Right:} Average perplexity of Real-FIM-Eval-Add and Real-FIM-Eval-Edit (lower is better).
}\label{fig:fig_plot}
\end{center}
\vskip -0.2in
\end{figure}

%% file: contents/related_work.tex
\section{Related Work}\label{sec:related_work}


\paragraph{LLMs for Code.}
Decoder-only LLMs have overwhelming popularity due to their superior scalability and generation performance~\citep{gpt3}. Coding is one of the most impactful domains for LLMs. LLMs can handle coding tasks such as code generation~\citep{codex,mbpp}, real-time code suggestions~\citep{copilot_arena}, and editing~\citep{swebench}, with applications ranging from synthesizing executable functions to automating repetitive development workflows~\citep{copilot_study}. \sida{should be safe to remove this paragraph. already covered in intro and eval sections}


\paragraph{Fill-in-the-Middle (FIM) Pretraining.}
While encoder-decoder models naturally employ a FIM-like objective ~\citep{bert,t5},
decoder-only models are trained to FIM via data transformation and FIM training became a defining feature of Code LLMs. Seminal work \citet{humaneval_fim} claimed the \textit{FIM-for-free} property: FIM rates up to 0.9 did not harm left-to-right (L2R) generation. But subsequent models like StarCoder~\citep{starcoder} and DeepSeek-Coder~\citep{deepseekcoder} cap FIM rates at 0.5. The DeepSeek-Coder technical report describes how higher FIM rate causes significant performance drop.
Our experiments confirm that traditional FIM with a 0.7 rate indeed degrades L2R performance, highlighting the need for methods that reconcile this tradeoff.

    
\paragraph{FIM Span Selection.}
While existing FIM methods uniformly sample spans at random, prior work in both natural language processing~\citep{spanbert,lm_fill_blank} and programming languages shows that selecting semantically or syntactically meaningful spans improves downstream task performance of trained models.
\cite{humaneval_fim} identifies the inability to properly connect to the suffix as a difficulty of FIM and \cite{horizonlength} show that length is important.
For code, spans often align with subtrees in Abstract Syntax Trees (ASTs): AST-T5~\citep{astt5} enhances T5's span corruption with syntax-aware pretraining, and Qwen2.5-Coder~\citep{qwencoder} uses AST-guided FIM during post-training. Using language-agnotic strategies to mask diverse AST structures, we are the first to develop effective AST-aware FIM for pretraining decoder-only models. 


\paragraph{Evaluation of Code FIM.}
Existing FIM benchmarks, including HumanEval Single-Line infilling~\citep{humaneval_fim}, SAFIM~\citep{safim}, and CruxEval~\citep{cruxeval}, suffer from artificial edit patterns that can misalign with real-world distributions.
Realistic benchmarks ~\citep{copilot_arena, swebench, repobench} are too noisy, expensive, or hard for base model development.
Optimizing for realism and high signal-to-noise ratio~\citep{ppl_noise, evalarena}, we propose Real-FIM-Eval, a benchmark that samples masks from real git commit patches to mirror real code edits. 

%% file: contents/method.tex
\section{Method}\label{sec:method}

We now describe our pretraining methods. We first review autoregressive language model (LM) pretraining and traditional random-character Fill-in-the-Middle (FIM) pretraining. Then, we introduce our core contribution, AST-Aware Fill-in-the-Middle (AST-FIM). This approach enhances FIM by using the code's Abstract Syntax Tree (AST) structure to generate training inputs from code data.

\subsection{Language Model Pretraining}\label{sec:ordinary_l2r}

LM pretraining starts by processing input text. The text is first tokenized into a sequence of subword tokens using Byte Pair Encoding (BPE). These tokens are then grouped into sequences of a fixed length, for example, 8,192 tokens in our models. The model is pretrained by predicting the next token in the sequence given the tokens that came before it. The negative log-likelihood (NLL) loss on all tokens is minimized. We call this standard method left-to-right (L2R) training.


\subsection{Fill-in-the-Middle (FIM) Pretraining}\label{sec:method_fim_pretraining}

FIM pretraining helps autoregressive decoder-only language models perform infilling tasks---generating text at a specific point within a given prompt conditioning on both the text before the point (\textit{prefix}) and the text after the point (\textit{suffix}) to generate the missing \textit{middle} part.



\paragraph{FIM Implementation.} Following \cite{humaneval_fim}, FIM works by data transformation and reduction to L2R training. First, a document is randomly split into a \textit{prefix}, a \textit{middle}, and a \textit{suffix}. Second, these parts are rearranged so the middle part comes last. Special sentinel tokens are used to combine these parts into a single sequence for the model. Common formats include:

\begin{itemize}
    \item \textit{Prefix-Suffix-Middle (PSM):} \texttt{[PRE] prefix [SUF] suffix [MID] middle [EOT]}
    \item \textit{Suffix-Prefix-Middle (SPM):} \texttt{[PRE] [SUF] suffix [MID] prefix middle [EOT]}
\end{itemize}

After that, the language model is trained on this rearranged sequence using NLL loss, just like in ordinary left-to-right training.


\paragraph{Joint Training with L2R.} FIM pretraining is usually done jointly with ordinary left-to-right (L2R) pretraining. During training, for each code example, the FIM transformation as described above is applied with a certain probability, called the \textbf{FIM rate} \(p\). With probability \(1 - p\), ordinary L2R training is used for this step,
where the entire code example is fed into the model using the method described in \Cref{sec:ordinary_l2r}.
When \(p\) is not too large, LMs trained this way can gain the infilling capability while keeping their L2R generation capability~\citep{humaneval_fim}.

\subsection{AST-Aware Fill-in-the-Middle}\label{sec:ast_fim_mask_strategy}


In FIM pretraining, each input document is divided into three parts: a prefix, a middle, and a suffix. Our approach introduces a key difference in how the middle section is selected, i.e., the \textbf{masking} algorithm. The traditional method, \textbf{Rand-FIM}, treats code as a sequence of characters. It selects the masked part by choosing character positions~\citep{codellama,humaneval_fim} or token positions~\citep{incoder} uniformly at random. In contrast, our method, \textbf{AST-Aware FIM (AST-FIM)}, selects the masked part based on the code's syntactical structure. Specifically, the masked part {\em always corresponds to one or more complete subtrees within the code's Abstract Syntax Tree (AST).} This way, AST-FIM respects the syntactical structure of the code, which gives the model a strong prior that matches how developers edit and complete code in practice.

\input{figures/fig_method}


\paragraph{Parsing code into ASTs.} To implement AST-FIM, we first parse the code into ASTs. We assume that the code files in our training data are syntactically correct. This assumption is generally safe, as we observed that 98.2\% code pushed to GitHub (our training dataset) is syntactically valid. \sida{since treesitter is supposed to handle errors, worth saying if you discard the errors?} \linyuan{In the ``FIM'' paragraph, we said ``we use Rand-FIM for programming languages not supported by Tree-sitter and for any code files that fail to parse correctly''.}
\linyuan{98.2\% of code can be parsed without errors by Tree-Sitter; but it doesn't mean that they can be compiled successfully using a compiler in that programming language. Shall we explain this point here or in the footnote?}\sida{i think good to explain, added to the CFA, execution section}
\alvin{say something about how the repos are selected or add future reference}\linyuan{Unfortunately the training dataset "GitHub Neo v3" was a Meta internal dataset. I am not sure how they scraped and processed the dataset.}
Our approach only requires parsing which is easier and more scalable than static analysis like Control-Flow Analysis (CFA), compilation, or code execution. For parsing, we use Tree-sitter~\citep{treesitter_2025}, a multi-language parser generator, to construct the ASTs\footnote{Tree-sitter produces Concrete Syntax Trees (CSTs), but the distinctions are not important for our purposes.}, where each subtree represents a consecutive span of characters in the code.


\paragraph{Masking Algorithm.} The FIM masking algorithm determines which span of code to designate as the ``middle'' part. The input to this algorithm is a code file. The algorithm then samples a span of code, ensuring that in AST-FIM, this sampled span aligns with the boundaries of AST subtrees. We designed our masking algorithm with two main goals:

\begin{enumerate}[leftmargin=*]
\item 
The algorithm should generate spans covering diverse AST node types to improve the model's generalization capability across various scenarios.
\item The algorithm must be language-agnostic. It should work on any AST without language-specific rules, which is important given that there are 100+ programming languages. 
\end{enumerate}

To achieve these goals, we use an equal mix of two masking methods: \textbf{Single-Node Masking} and \textbf{Aligned-Span Masking}.


\paragraph{Single-Node Masking.} This algorithm selects a single AST node to mask. We sample an AST node with probability proportional to its size (number of bytes in its corresponding text). Masked nodes typically include binary expressions, call expressions, if statements, function definitions, and so on. We ignore nodes that represent keywords or punctuations, as these are not typically considered part of the \textit{abstract} syntax tree, even though Tree-sitter often parse them as leaf nodes.





\paragraph{Aligned Span Masking.}
Single-node masking selects only a single AST subtree. But code edits often involve multiple adjacent elements, like multiple statements or multiple methods within a class. Aligned Span Masking allows selecting multiple adjacent AST nodes as follows:

\begin{enumerate}[leftmargin=*]
\item Sample a character span with uniformly random endpoints \texttt{[start, end]}.
\item Find the \textit{lowest}\footnote{By ``low'' in tree we mean ``far from the root node'', following the standard terminology used in graph theories, e.g.,, in \url{https://en.wikipedia.org/wiki/Lowest_common_ancestor}} AST subtree $T$ that contains this character span. 
\item Let $T_1, T_2, \dots, T_n$ be the direct children of $T$, the subtree selected in the previous step. We now select a continuous sequence of these children, $T_i, T_{i+1}, \dots, T_j$, to form the middle part. We choose the sequence that has the largest character-level \textit{Intersection over Union (IoU)} with the original random span \texttt{[start, end]}.
\end{enumerate}


\paragraph{Training Process.} Single-Node Masking and Aligned Span Masking determine how we select the middle part for our AST-FIM pretraining. Then, we format the input as either the PSM or SPM with special tokens. The model is then trained to predict the middle part using NLL loss. Similar to Rand-FIM, our AST-FIM is typically performed jointly with ordinary L2R objective to ensure the model retains L2R generation capabilities while learning how to fill-in-the-middle.

%% file: figures/fig_method.tex
\begin{figure}[ht]
\vskip 0.2in
\begin{center}
\centerline{\includegraphics[scale=0.85]{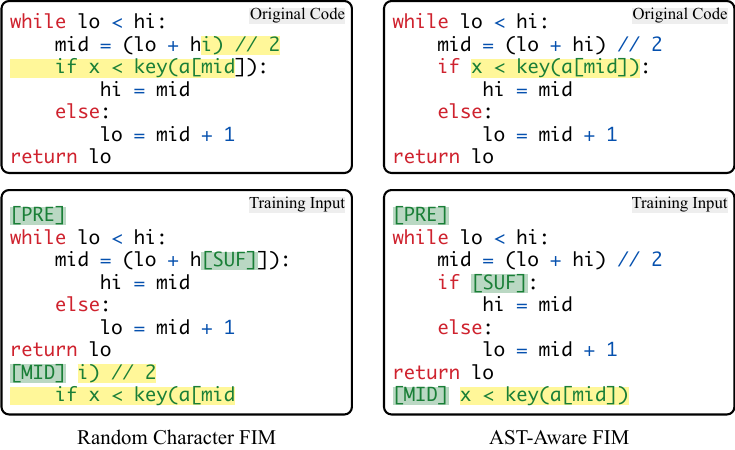}}
\caption{
\textbf{Comparison of training inputs processed by Rand-FIM and AST-FIM using the PSM format.} Given the same code, Rand-FIM selects a random character span as the ``middle'' part, while AST-FIM selects a span corresponding to entire AST subtrees. AST-FIM generates cleaner training examples that better reflect practical code completion scenarios.
}\label{fig:fig_method}
\end{center}
\vskip -0.4in
\end{figure}

%% file: contents/real_fim_eval_benchmark.tex
\section{The Real-FIM-Eval Benchmark}\label{sec:real_fim_eval_benchmark}


We introduce a new benchmark called Real-FIM-Eval to evaluate FIM capabilities in scenarios that reflect real-world code completion based on GitHub commits. The dataset is released here: \url{https://huggingface.co/datasets/gonglinyuan/real_fim_eval}

\subsection{Benchmark Construction}\label{sec:real_fim_eval_benchmark_construction}

\paragraph{Data Source.} Our benchmark is built using data from recent GitHub commits between Jan. 2025 and Feb. 2025. These commits originate from 228 permissively licensed GitHub repositories with 10,000+ stars, spanning the top 12 widely-used programming languages. \Cref{app:stats_real_fim_eval} shows the distribution of examples across these languages. The data collection period is entirely separate from the data used for pretraining our models, minimizing the potential impact of data contamination.

\input{figures/fig_real_fim_eval}

\paragraph{Splits.} We process git commits using \texttt{diff\_match\_patch}\footnote{{\ttfamily \url{https://github.com/google/diff-match-patch}}} to identify line-level changes. The commits are then categorized into two  splits for Real-FIM-Eval, as visualized in \Cref{fig:fig_real_fim_eval}:

\begin{itemize}[leftmargin=*]
    \item \textbf{Add} (17,879 examples): This split uses git commits where a developer added a new segment of code into an existing file. To create the FIM prompt, we treat the added code segment as the ``middle'' part that the language model needs to predict. The code surrounding the addition forms the prefix (code before) and the suffix (code after).
    \item \textbf{Edit} (13,922 examples): This split uses git commits where a developer modified existing code by removing a segment and replacing it with a new one. We present this task to LLMs in a \textit{conflict-merge} format. The prompt includes the code context (prefix and suffix) and marks the original code segment (to be removed). The model is asked to infill the updated code segment.
\end{itemize}

\subsection{Evaluation Metric}

We evaluate model performance on Real-FIM-Eval using character-level perplexity. This metric measures how well the model predicts the sequence of characters in the ground truth ``middle'' part. The perplexity is calculated as
$ \exp\left(-\frac{1}{\mathrm{n\_chars}(\mathrm{mid})} \sum_{i\in\mathrm{mid}} \log p_{i}\right)$.
In this formula, \(p_{i}\) represents the probability the model assigns to the \(i\)-th token of the ground truth (only the ``middle'' part that the model is tasked to infill), and $\mathrm{n\_chars}(\mathrm{mid})$ is the total number of characters in that ground truth middle segment.

We use perplexity for evaluation as it provides a scalable and low-noise signal, which helps in obtaining stable comparisons between different models~\citep{ppl_noise}. We do not use execution-based evaluation, as gathering and executing unit tests from arbitrary repos at scale is impractical. Future work could explore generative metrics such as exact match or edit distance.

%% file: figures/fig_real_fim_eval.tex
\begin{figure}[tb]
\vskip 0.2in
\begin{center}
\centerline{\includegraphics[scale=0.85]{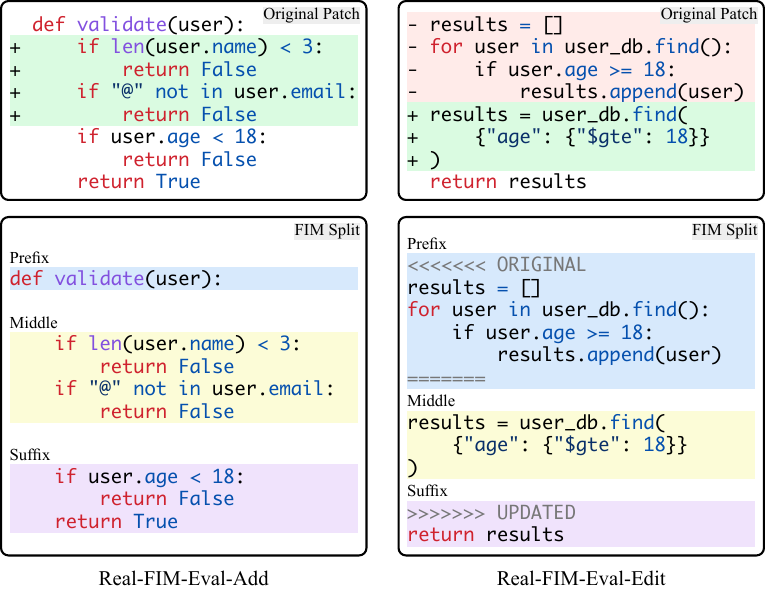}}
\caption{
\textbf{Construction of Fill-in-the-Middle (FIM) examples for the proposed Real-FIM-Eval benchmark splits, derived from real-world git commits.} \textbf{Add:} Uses code insertions; the added code becomes the ``middle'' to predict. \textbf{Edit:} Uses code modifications, presented via a conflict-merge format contrasting the \texttt{ORIGINAL} and \texttt{UPDATED} code within the surrounding context. The content of the added code is the ``middle'' part to predict.
}\label{fig:fig_real_fim_eval}
\end{center}
\vskip -0.2in
\end{figure}

%% file: contents/experimental_setup.tex
\section{Experimental Setup}\label{sec:experimental_setup}

This section outlines our experimental setup. We train 1B and 8B models using Rand-FIM and AST-FIM. We evaluate these models on standard code generation tasks and FIM tasks, including Real-FIM-Eval. The goal is to test if our AST-FIM models outperform other FIM methods on code completion while performing comparably to standard L2R models on standard generation tasks.

\subsection{Training}\label{sec:experimental_setup_training}






\paragraph{Model.} We conduct experiments using Llama-3-1B and Llama-3-8B model architectures, training each model from scratch with random initializations.  
For tokenization, we use the Llama-3 version of \texttt{tiktoken cl100k\_base}\footnote{{\ttfamily \url{https://github.com/openai/tiktoken}}} tokenizer extending the the vocabulary to 128k.

\paragraph{Data.} Following recent code models, our training data contains 90\% programming language data from GitHub and 10\% natural language data. The code data is sourced from GitHub repositories and decontaminated for evaluation data such as HumanEval and MBPP. The natural language data includes 8\% from Apple DCLM~\citep{dclm} and 2\% from Wikipedia. We use Tree-sitter to parse the code data.

\paragraph{FIM.} we use context-level FIM instead of document-level FIM~\citep{humaneval_fim}. We train the models jointly with 70\% FIM objectives and 30\% L2R objectives. Within the FIM portion, 90\% use our AST-FIM, and 10\% use Rand-FIM. \linyuan{Added:} For both AST-FIM and Rand-FIM, we use 50\% PSM and 50\% SPM mode. We also use Rand-FIM for programming languages not supported by Tree-sitter and for any code files that fail to parse correctly. For the natural language data, we only apply the standard L2R objective without any FIM. 

\paragraph{Hyperparameters.} We train the models using 256 H100 GPUs. We train 1B models for 500B tokens. This corresponded to 120,000 steps, using a context window of 8192 tokens. The batch size is 2 contexts per GPU across the 256 GPUs. For 8B models, we train 2$\times$ or 4$\times$ more steps, corresponding to 1T or 2T tokens. See \Cref{app:hyperparameters} for more details about hyperparameters.

\subsection{Evaluation}


We evaluate models on two types of tasks: ordinary L2R generation tasks and FIM tasks. Our goal is to evaluate: \textbf{(a)} if models trained with our proposed AST-FIM method perform almost as well as standard L2R models on L2R tasks; \textbf{(b)} if AST-FIM models perform better than models trained with Rand-FIM method on FIM tasks in real-world scenarios.


\paragraph{L2R Tasks.} We use the HumanEval+ and MBPP+ text-to-code generation benchmarks~\citep{evalplus} to evaluate. Following the original MBPP paper~\citep{mbpp}, we use a standard 3-shot setup for evaluating on MBPP+. We measure performance using the Pass@1 metric for both benchmarks. Pass@1 indicates the percentage of examples for which the model generates functionally correct code in a single attempt, evaluated using test cases.


\paragraph{FIM tasks.} We use the SAFIM~\citep{safim} and Real-FIM-Eval for FIM evaluation. SAFIM tests how well models can infill different types of code structures given both text and code context. SAFIM uses Pass@1 as the metric. Details of Real-FIM-Eval can be found in \Cref{sec:real_fim_eval_benchmark}.

When evaluating all FIM models, we use the PSM prompt format. We also evaluate L2R models on FIM tasks, even though standard L2R models typically perform poorly on FIM tasks. We try our best to get reasonable results by using the prompt format detailed in \Cref{app:l2r_prompt}.

  
We evaluate the pretrained base models directly without any further training or fine-tuning. For all generation tasks, both L2R and FIM, we use greedy decoding to generate the code.

%% file: contents/evaluation_results.tex
\section{Evaluation Results}\label{sec:evaluation_results}

This section presents evaluation results comparing AST-FIM against the Rand-FIM baseline and other off-the-shelf code LLMs.

\subsection{Comparison of Pretraining Methods}


We compare our AST-FIM pretraining method against two baselines: random character FIM (Rand-FIM) and ordinary L2R pretraining. To ensure a fair comparison, all models use the same Llama-3 1B architecture, same datasets, and the same computational environments. The only difference lies in the pretraining objective applied. The evaluation results comparing these three methods are presented in \Cref{tab:main_1}. \Cref{fig:fig_plot} tracks the performance of each model on FIM tasks across checkpoints during pretraining.

\input{tables/tab_main_1}


\paragraph{AST-FIM Improves LLM's Capability in Practical FIM tasks.}
The results shown in \Cref{tab:main_1} show that AST-FIM outperforms the Rand-FIM baseline across all subtasks in the SAFIM benchmark. This result is expected, as AST-FIM directly trains the model to complete masked AST structures, which aligns well with the SAFIM benchmark's objective of evaluating AST structure completion. More importantly, AST-FIM achieves better performance than Rand-FIM on the Real-FIM-Eval benchmark.
Compared to Rand-FIM, the training signal provided by AST-FIM is better aligned with real-world code editing patterns, leading to superior performance on realistic FIM tasks.
Furthermore, \Cref{fig:fig_plot} shows this performance advantage is consistent throughout training.
With models trained less than a single epoch, \Cref{fig:fig_plot} also highlights AST-FIM's data efficiency: AST-FIM reaches performance similar to Rand-FIM after seeing only 50-70\% of the training tokens. 


\paragraph{AST-FIM Retains L2R Capability.}
Training models with a high FIM rate can negatively impact their performance on standard L2R code generation tasks. While the optimal FIM rate is yet to be determined (with suggestions ranging from 0.5 to 0.9 in different contexts~\citep{starcoder,deepseekcoder,codellama,humaneval_fim}), we used a FIM rate of 0.7 in our experiments. \Cref{tab:main_1} shows that the Rand-FIM model, trained at this rate, experiences a significant performance decrease on the L2R benchmarks HumanEval+ and MBPP+ compared to the ordinary L2R model. One possible reason is that character-level random FIM can break coherent code structures and introduce noisy boundaries between the prefix, suffix, and middle parts, potentially harming the model's understanding of standard code generation. In contrast, the AST-FIM model achieves L2R performance nearly identical to the baseline L2R model on both HumanEval+ and MBPP+. This suggests that by operating on meaningful code structures, AST-FIM preserves the structural coherence of code, similar to standard L2R training.
As further evidence, the AST-FIM training loss is at the middle of the higher Rand-FIM loss and the lower L2R loss throughout training, indicating that AST-FIM is easier than rand-FIM but harder than L2R.
As a result, AST-FIM enables the integration of strong FIM capabilities into a language model with minimal impact on its L2R performance.

\subsection{Evaluating AST-FIM at Scale}


We next evaluate AST-FIM at a larger scale. We train an 8B AST-FIM model for 1T and 2T tokens respectively, and a Rand-FIM model for 1T tokens with other settings being identical as our earlier model. The AST-FIM 8B (2T) model is compared to other publicly available \textit{base} (pretrain-only, non-finetuned, non-distilled) models of similar sizes. AST-FIM-8B (2T) and L2R-8B (8T) are both trained on the same 2T code tokens over 2 epochs. The total training tokens reported could be over multiple epochs.  The results of these experiments are shown in \Cref{tab:main_2}.

\input{tables/tab_main_2}


\paragraph{AST-FIM Method Can Scale.}
The findings from our 1B scale experiments remain consistent at the 8B scale. The AST-FIM model with 1T tokens outperforms the Rand-FIM model trained under the same conditions. This improvement is observed in both standard L2R tasks (HumanEval+ and MBPP+) and FIM tasks (SAFIM and Real-FIM-Eval).
\linyuan{Added a conclusion sentence:} This consistency across scales highlights the scalability of the AST-FIM pretraining method.
\alvin{I don't get this point, do you mean both Rand and AST are trained using multiple epochs but AST is better?} \linyuan{Removed this point. Sida's point was that:
Rand-FIM has largest randomness
-> Rand-FIM is least likely to overfit
-> Rand-FIM is data efficient;
We are better than Rand-FIM using less randomness
-> Our method itself is more data efficient;
I think it makes some sense to me, but the logic is perhaps too complicated. Maybe let's just remove it, because we already claimed data efficiency using a simpler argument in Sec 6.1.
}\sida{ok, agree. I just added some info about multi epoch without commenting}


\paragraph{Performance of AST-FIM-8B is Competitive Against Similar-Sized Models.}
Our AST-FIM model trained on 2T tokens shows competitive performance against other base models with 6B to 8B parameters. Notably, AST-FIM achieves significantly better results than most of the compared models on both FIM benchmarks, SAFIM and Real-FIM-Eval. 

%% file: tables/tab_main_1.tex
\begin{table}[t]
\caption{
\textbf{Comparison of 1B-parameter code LLMs}. All the models are trained under identical conditions. We evaluate FIM models using PSM prompt, and L2R models using a SPM prompt without special tokens (See \Cref{app:l2r_prompt}).
}
\label{tab:main_1}
\begin{center}
\scalebox{0.9}{
\begin{tabular}{lccccccc}
\toprule
& \textbf{HumanEval+} & \textbf{MBPP+} & \textbf{SAFIM} & \textbf{SAFIM} & \textbf{SAFIM} & \textbf{RealFIM} & \textbf{RealFIM} \\
& & & \small{\textbf{Algorithm}} & \small{\textbf{Control}} & \small{\textbf{API}} & \small{\textbf{Add}} & \small{\textbf{Edit}} \\
& \small{Pass@1$\uparrow$} & \small{Pass@1$\uparrow$} & \small{Pass@1$\uparrow$} & \small{Pass@1$\uparrow$} & \small{Pass@1$\downarrow$} & \small{PPL$\downarrow$} & \small{PPL$\downarrow$}  \\ \midrule
L2R & 15.9 & 34.5 & 26.5 & 31.0 & 48.1 & 1.390 & 1.286 \\
Rand-FIM & 11.6 & 31.5 & 28.2 & 36.1 & 56.1 & 1.283 & 1.232 \\
\textbf{AST-FIM} & 15.9 & 35.3 & \textbf{33.5} & \textbf{41.2} & \textbf{60.3} & \textbf{1.269} & \textbf{1.230} \\
\bottomrule
\end{tabular}
}
\end{center}
\end{table}

%% file: tables/tab_main_2.tex
\begin{table}[t]
\caption{
\textbf{Comparison of 6B-8B parameter code LLMs.}
\textbf{Top:} AST-FIM-8B (1T) vs. Rand-FIM-8B (1T), trained under identical conditions.
\textbf{Bottom:} AST-FIM-8B (2T) vs. L2R-8B (8T) and publicly available \emph{base} models. The pretraining token count of each model is given in the parenthesis with + for continued pretraining. L2R-8B (8T), Llama-3.1 and our models are evaluated using our codebase; for other models, we use Huggingface Transformers to evaluate them on FIM tasks. For HumanEval+ and MBPP+, we use their reported numbers or those reported by EvalPlus ~\citep{evalplus}
}
\label{tab:main_2}
\begin{center}
\scalebox{0.81}{
\begin{tabular}{lccccccc}
\toprule
& \textbf{HumanEval+} & \textbf{MBPP+} & \textbf{SAFIM} & \textbf{SAFIM} & \textbf{SAFIM} & \textbf{RealFIM} & \textbf{RealFIM} \\
\small{Subset} & & & \small{\textbf{Algorithm}} & \small{\textbf{Control}} & \small{\textbf{API}} & \small{\textbf{Add}} & \small{\textbf{Edit}} \\
\small{Metric} & \small{Pass@1$\uparrow$} & \small{Pass@1$\uparrow$} & \small{Pass@1$\uparrow$} & \small{Pass@1$\uparrow$} & \small{Pass@1$\uparrow$} & \small{PPL$\downarrow$} & \small{PPL$\downarrow$}  \\
\small{Num Examples} & \small{164} & \small{371} & \small{8,731} & \small{8,629} & \small{310} & \small{17,879} & \small{13,922} \\
\midrule
Rand-FIM-8B (1T) & 32.3 & 59.8 & 50.0 & 56.5 & 62.9 & 1.225 & 1.172 \\
\textbf{AST-FIM-8B (1T)} & \textbf{37.8} & \textbf{63.6} & \textbf{55.0} & \textbf{61.7} & \textbf{70.3} & \textbf{1.215} & \textbf{1.164} \\ \midrule
Llama-3.1-8B (15T) & 29.9 & 51.5 & 39.9 & 43.0 & 48.1 & 1.337 & 1.264 \\
L2R-8B (8T) & 34.1 & 51.2 & 42.5 & 45.3 & 53.5 & 1.340 & 1.253 \\
CodeLlama-7B (+1T) & 35.4 & 37.8 & 34.7 & 53.6 & 46.8 & 1.222 & 1.175 \\
StarCoder2-7B (4.3T) & 29.9 & 35.4 & 46.2 & 58.4 & 70.6 & 1.227 & 1.184 \\
CodeGemma-7B (+1T) & 41.5 & 44.5 & 50.8 & 65.4 & 73.5 & 1.217 & 1.187 \\
DeepSeek-Coder-6.7B (2T) & 39.6 & 47.6 & 54.7 & \textbf{65.8} & 69.7 & 1.254 & 1.206 \\
Qwen2.5-Coder-7B (+5.5T) & \textbf{53.0} & 62.9 & 53.0 & 59.2 & 73.9 & 1.212 & 1.167 \\
\textbf{AST-FIM-8B (2T)} & 42.1 & \textbf{65.2} & \textbf{57.0} & 63.2 & \textbf{74.5} & \textbf{1.210} & \textbf{1.160} \\
\bottomrule
\end{tabular}
}
\end{center}
\end{table}

%% file: contents/limitations.tex
\section{Limitations}\label{sec:limitations}

\linyuan{TODO: Put into appendix if we don't have space.}\sida{there should be enough space. let me try merging limitation into discussions}\linyuan{I have a cross-reference to this section in the paper checklist; please update it accordingly if this is moved to somewhere else.}


\paragraph{L2R.}
AST-FIM did not show clear improvements on left-to-right (L2R) generation tasks when compared to models trained solely for L2R. This is consistent with other works in decoder-only FIM.
Training using AST-FIM is better on L2R tasks compared to training using Rand-FIM.
Outperforming standard L2R training on L2R tasks would be the strongest evidence for the benefits of syntax annotations. 
\paragraph{Random FIM.}
On the HumanEval random span FIM task~\citep{santacoder}, AST-FIM is slightly worse than Rand-FIM, which has a better matching training distribution.
With 10\% random FIM in the data mix, the AST-FIM model reached the same performance as the Rand-FIM model on random span FIM by the end of training but got there slower. Without the 10\% random FIM data, the model will learn the deterministic patterns of AST and fail at random FIM. Better matching training and testing distributions still result in stronger performance for various FIM tasks and offer a good explanation why AST-FIM is strong on SAFIM and Real-FIM. A more careful test of random FIM, token FIM, and line FIM with tuned lengths is needed to better understand the effects of mid length.



%% file: contents/conclusion.tex
\section{Conclusion}\label{sec:conclusion}

Traditional random-character FIM pretraining is suboptimal for real-world applications due to a mismatch with how developers edit code. AST-FIM addresses this limitation by aligning the pretraining objective with code's syntactic structure through AST-aware subtree masking, enabling models to learn more realistic infilling patterns. The Real-FIM-Eval benchmark, based on real-world GitHub commits, validates AST-FIM's strengths in practical code completion and editing tasks. Across model scales, AST-FIM consistently outperforms Rand-FIM, showing scalability and generalizability across programming languages. 


%% file: contents/acknowledgement.tex
\section*{Acknowledgements}
We thank Gabriel Synnaeve and Jade Copet for support of the project and enabling a strong baseline to build on; Zach Rait for discussions on how to evaluate FIM for code completion models; Jade Copet for help with data processing and evaluation; Baptiste Rozi\`{e}re and Fabian Gloeckle for helpful discussions about context-level FIM; Yuxiang Wei for showing us how to work with patches.


%% file: contents/appendix_l2r_prompt.tex
\subsection{Prompting L2R Models for FIM Evaluation}\label{app:l2r_prompt}


Evaluating L2R models on FIM tasks is a challenge. Although L2R models typically perform poorly on FIM tasks, we try to obtain the most reasonable results possible.

The challenge is that L2R models do not recognize the special tokens, such as \texttt{[PRE]}, \texttt{[SUF]}, and \texttt{[MID]}, which are essential for the PSM or SPM prompt formats described in \Cref{sec:method_fim_pretraining}. So we could not use those prompts directly for L2R models.

To address this, we experimented with two alternative prompt formats for L2R models on FIM tasks:

\begin{enumerate}
    \item \textbf{Prefix-only:} In this format, we only provide the prefix to the model. The model's task is to continue generating code, effectively completing the missing part after the prefix.
    \item \textbf{SPM without special tokens:} This format mimics the structure of SPM but avoids special tokens. The input prompt is structured as: \texttt{suffix} $\dlsh\dlsh$ \texttt{prefix middle}
    
    We use two line breaks to separate the suffix and the prefix. The model is expected to generate the missing middle part immediately following the prefix in the input.
\end{enumerate}

We find that the ``SPM without special tokens'' prompt consistently produced better results. This is because providing the suffix offers valuable context that the L2R model can leverage, even though it is not explicitly trained on FIM tasks with special tokens. Therefore, for all FIM task evaluations involving L2R models presented in this paper, we use the ``SPM without special tokens'' prompt format.

%% file: contents/appendix_stats_real_fim_eval.tex
\subsection{Statistics about Real-FIM-Eval}\label{app:stats_real_fim_eval}

\input{tables/tab_real_fim_eval_stats}

\Cref{tab:real_fim_eval_stats} shows the distribution of examples across the 12 languages in the Real-FIM-Eval benchmark.

%% file: tables/tab_real_fim_eval_stats.tex
\begin{table}[t]
\caption{
\textbf{Distribution of programming languages} in the Real-FIM-Eval benchmark.
}
\label{tab:real_fim_eval_stats}
\begin{center}
\scalebox{0.82}{
\begin{tabular}{lccccccccccc}
\toprule
\textbf{Python} & \textbf{Rust} & \textbf{Java} & \textbf{C++} & \textbf{TypeScript} & \textbf{Go} & \textbf{Ruby} & \textbf{C\#} & \textbf{JavaScript} & \textbf{Kotlin} & \textbf{PHP} & \textbf{Scala} \\ \midrule
6,271 & 4,727 & 3,716 & 3,265 & 3,182 & 2,587 & 1,686 & 1,563 & 1,502 & 1,440 & 1,396 & 466 \\
\bottomrule
\end{tabular}
}
\end{center}
\end{table}

%% file: contents/appendix_hyperparameters.tex
\subsection{Prompting L2R Models for FIM Evaluation}\label{app:hyperparameters}

\input{tables/tab_hyperparameter}

The pretraining hyperparameter we used to train AST-FIM models, as well as L2R and Rand-FIM baselines, are detailed in \Cref{tab:appendix_hyperparameter}.

%% file: tables/tab_hyperparameter.tex
\begin{table}[t]
\caption{\textbf{Pretraining hyperparameters for AST-FIM models.}}
\label{tab:appendix_hyperparameter}
\begin{center}
\begin{tabular}{lrrr}
\toprule
& \textbf{1B} & \textbf{8B/1T} & \textbf{8B/2T} \\ \midrule
Model Architecture & Llama-3-1B & Llama-3-8B & Llama-3-8B \\
Optimizer & AdamW & AdamW & AdamW \\
Peak Learning Rate & 5e-4 & 1e-3 & 1e-3 \\
Learning Rate Schedule & Cosine & Cosine & Cosine \\
Adam \(\beta_1\) & 0.9 & 0.9 & 0.9 \\
Adam \(\beta_2\) & 0.95 & 0.95 & 0.95 \\
Gradient Clip Norm & 1.0 & 1.0 & 1.0 \\
Warm-Up Steps & 2,000 & 2,000 & 2,000  \\
Total Steps & 120,000 & 240,000 & 480,000 \\
Context Window (Tokens) & 8,192 & 8,192 & 8,192 \\
Total Batch Size & 512 & 512 & 512 \\
\bottomrule
\end{tabular}
\end{center}
\end{table}

%% file: contents/appendix_method_analysis.tex
\subsection{Analysis of AST-FIM Masking Strategies}\label{app:method_analysis}

As described in \Cref{sec:ast_fim_mask_strategy}, AST-FIM uses two masking strategies: Single-Node Masking and Aligned-Span Masking. To evaluate their individual contributions, we conduct an ablation study. The results are shown in \Cref{tab:method_analysis}. The models for this experiment are trained at 1B scale (same as \Cref{tab:main_1}), using either Single-Node Masking or Aligned-Span Masking only.

\input{tables/tab_method_analysis}

The result shows that Single-Node Masking outperforms Aligned-Span Masking on SAFIM-Control, SAFIM-API, because these tasks involve completing structures that  correspond to single AST nodes, aligning well with this masking approach. Single-Node Masking also yields better results on Real-FIM-Eval-Add, because most real-world code insertions involves adding individual AST nodes. On other tasks, the performance differences between models are not statistically significant. Our final AST-FIM model incorporates a 50:50 combination of both strategies. This hybrid approach achieves slightly improved overall performance compared to using either Single-Node or Aligned-Span Masking in isolation, likely benefiting from the diverse structural priors each strategy provides.

%% file: tables/tab_method_analysis.tex
\begin{table}[t]
\caption{
\textbf{Comparison of AST-FIM masking strategies.} We compare the performance of 1B models trained exclusively with either Single-Node Masking or Aligned-Span Masking, evaluated on various code completion benchmarks.
}
\label{tab:method_analysis}
\begin{center}
\scalebox{0.82}{
\begin{tabular}{lccccccc}
\toprule
& \textbf{HumanEval+} & \textbf{MBPP+} & \textbf{SAFIM} & \textbf{SAFIM} & \textbf{SAFIM} & \textbf{RealFIM} & \textbf{RealFIM} \\
& & & \small{\textbf{Algorithm}} & \small{\textbf{Control}} & \small{\textbf{API}} & \small{\textbf{Add}} & \small{\textbf{Edit}} \\
& \small{Pass@1$\uparrow$} & \small{Pass@1$\uparrow$} & \small{Pass@1$\uparrow$} & \small{Pass@1$\uparrow$} & \small{Pass@1$\downarrow$} & \small{PPL$\downarrow$} & \small{PPL$\downarrow$}  \\ \midrule
L2R & 15.9 & 34.5 & 26.5 & 31.0 & 48.1 & 1.390 & 1.286 \\
Rand-FIM & 11.6 & 31.5 & 28.2 & 36.1 & 56.1 & 1.283 & 1.232 \\
AST-FIM (Single Node) & 15.9 & 35.0 & 34.6 & 39.0 & 57.1 & 1.272 & 1.229 \\
AST-FIM (Aligned Span) & \textbf{17.7} & 34.8 & \textbf{35.3} & 37.7 & 55.5 & 1.286 & \textbf{1.226} \\
AST-FIM (Both Methods) & 15.9 & \textbf{35.3} & 33.5 & \textbf{41.2} & \textbf{60.3} & \textbf{1.269} & 1.230 \\
\bottomrule
\end{tabular}
}
\end{center}
\end{table}

%% file: contents/appendix_broader_impact.tex
\subsection{Broader Impact}\label{app:broader_impact}

In this paper, we introduce AST-FIM, a Large Language Model (LLM) for code generation. The advancement of LLMs in code generation raises concerns about automated code production's security, privacy, and potential misuse. There is a risk that improved code generation capabilities could be exploited for malicious purposes, such as automating the creation of software vulnerabilities or facilitating the development of harmful software. Our research emphasizes the importance of responsible AI development and use, advocating for continuous monitoring, ethical guidelines, and safeguards to mitigate these risks.

%% file: contents/paper_checklist.tex
\newpage
\section*{NeurIPS Paper Checklist}

\begin{enumerate}

\item {\bf Claims}
    \item[] Question: Do the main claims made in the abstract and introduction accurately reflect the paper's contributions and scope?
    \item[] Answer: \answerYes{} 
    \item[] Justification: The contribution is summarized in the last 3 paragraphs of the introduction.

\item {\bf Limitations}
    \item[] Question: Does the paper discuss the limitations of the work performed by the authors?
    \item[] Answer: \answerYes{} 
    \item[] Justification: The limitations are discussed in \Cref{sec:limitations}.

\item {\bf Theory assumptions and proofs}
    \item[] Question: For each theoretical result, does the paper provide the full set of assumptions and a complete (and correct) proof?
    \item[] Answer: \answerNA{} 
    \item[] Justification: This paper does not include theoretical results.

    \item {\bf Experimental result reproducibility}
    \item[] Question: Does the paper fully disclose all the information needed to reproduce the main experimental results of the paper to the extent that it affects the main claims and/or conclusions of the paper (regardless of whether the code and data are provided or not)?
    \item[] Answer: \answerYes{} 
    \item[] Justification: A detailed description of our method is provided in \Cref{sec:method}. For evaluation, we use publicly available benchmarks, including Real-FIM-Eval which is included in the supplementary material.

\item {\bf Open access to data and code}
    \item[] Question: Does the paper provide open access to the data and code, with sufficient instructions to faithfully reproduce the main experimental results, as described in supplemental material?
    \item[] Answer: \answerYes{} 
    \item[] Justification: The data and code are included in the supplementary material.

\item {\bf Experimental setting/details}
    \item[] Question: Does the paper specify all the training and test details (e.g., data splits, hyperparameters, how they were chosen, type of optimizer, etc.) necessary to understand the results?
    \item[] Answer: \answerYes{} 
    \item[] Justification: Hyperparameters are detailed in \Cref{app:hyperparameters}.

\item {\bf Experiment statistical significance}
    \item[] Question: Does the paper report error bars suitably and correctly defined or other appropriate information about the statistical significance of the experiments?
    \item[] Answer: \answerYes{} 
    \item[] Justification: We do not plot error bars because pretraining LLMs with multiple random seeds is too expensive. Instead, in \Cref{fig:fig_plot}, we include compare the performance of different models for each checkpoint, as an alternative way to show statistical significance of the experiments.

\item {\bf Experiments compute resources}
    \item[] Question: For each experiment, does the paper provide sufficient information on the computer resources (type of compute workers, memory, time of execution) needed to reproduce the experiments?
    \item[] Answer: \answerYes{} 
    \item[] Justification: Those are detailed in \Cref{sec:experimental_setup}.
    
\item {\bf Code of ethics}
    \item[] Question: Does the research conducted in the paper conform, in every respect, with the NeurIPS Code of Ethics \url{https://neurips.cc/public/EthicsGuidelines}?
    \item[] Answer: \answerYes{} 
    \item[] Justification: The research conducted in the paper conform, in every respect, with the NeurIPS Code of Ethics.

\item {\bf Broader impacts}
    \item[] Question: Does the paper discuss both potential positive societal impacts and negative societal impacts of the work performed?
    \item[] Answer: \answerYes{} 
    \item[] Justification: The societal impacts are discussed in \Cref{app:broader_impact}.
    
\item {\bf Safeguards}
    \item[] Question: Does the paper describe safeguards that have been put in place for responsible release of data or models that have a high risk for misuse (e.g., pretrained language models, image generators, or scraped datasets)?
    \item[] Answer: \answerNA{} 
    \item[] Justification: We do not release the model.

\item {\bf Licenses for existing assets}
    \item[] Question: Are the creators or original owners of assets (e.g., code, data, models), used in the paper, properly credited and are the license and terms of use explicitly mentioned and properly respected?
    \item[] Answer: \answerYes{} 
    \item[] Justification: Discussed in \Cref{sec:experimental_setup_training} (for training data of AST-FIM) and \Cref{sec:real_fim_eval_benchmark_construction} (for the Real-FIM-Eval benchmark).

\item {\bf New assets}
    \item[] Question: Are new assets introduced in the paper well documented and is the documentation provided alongside the assets?
    \item[] Answer: \answerYes{} 
    \item[] Justification: The new asset (the Real-FIM-Eval benchmark) is detailed in \Cref{sec:real_fim_eval_benchmark_construction}.

\item {\bf Crowdsourcing and research with human subjects}
    \item[] Question: For crowdsourcing experiments and research with human subjects, does the paper include the full text of instructions given to participants and screenshots, if applicable, as well as details about compensation (if any)? 
    \item[] Answer: \answerNA{} 
    \item[] Justification: The paper does not involve crowdsourcing nor research with human subjects.

\item {\bf Institutional review board (IRB) approvals or equivalent for research with human subjects}
    \item[] Question: Does the paper describe potential risks incurred by study participants, whether such risks were disclosed to the subjects, and whether Institutional Review Board (IRB) approvals (or an equivalent approval/review based on the requirements of your country or institution) were obtained?
    \item[] Answer: \answerNA{} 
    \item[] Justification:  The paper does not involve crowdsourcing nor research with human subjects.

\item {\bf Declaration of LLM usage}
    \item[] Question: Does the paper describe the usage of LLMs if it is an important, original, or non-standard component of the core methods in this research? Note that if the LLM is used only for writing, editing, or formatting purposes and does not impact the core methodology, scientific rigorousness, or originality of the research, declaration is not required.
    \item[] Answer: \answerYes{} 
    \item[] Justification: This entire paper is about pretraining of LLMs.

\end{enumerate}